%% file: 0main_cvpr.tex
\documentclass[10pt,twocolumn,letterpaper]{article}

\usepackage[pagenumbers]{cvpr} 
\usepackage{microtype}
\usepackage{booktabs} 
\usepackage[numbers]{natbib}
\usepackage{graphicx}
\usepackage{amsmath}
\usepackage{amssymb}
\newcommand{\real}{\mathcal{R}}

\usepackage[pagebackref,breaklinks,colorlinks]{hyperref}
\usepackage{xcolor}

\usepackage[capitalize]{cleveref}
\crefname{section}{Sec.}{Secs.}
\Crefname{section}{Section}{Sections}
\Crefname{table}{Table}{Tables}
\crefname{table}{Tab.}{Tabs.}

\def\cvprPaperID{*****} 
\def\confName{CVPR}
\def\confYear{2023}
\definecolor{coolblack}{rgb}{0.0, 0.18, 0.39}
\definecolor{midnightblue}{rgb}{0.1, 0.1, 0.44}
\definecolor{darkmidnightblue}{rgb}{0.0, 0.2, 0.4}
\definecolor{palatinatepurple}{rgb}{0.41, 0.16, 0.38}

\begin{document}

\title{The Vanishing Decision Boundary Complexity and the Strong First Component}

\author{
Hengshuai Yao\\
Sony AI\\
University of Alberta\\
{\tt\small hengshu1@ualberta.ca}
}

\maketitle

\input{0abstract}

\input{0introduction}

\input{0db_evolve}

\input{0insights}


\input{0optimizer_sigma1}

\input{0model_compare}

\input{0related}

\input{0conclusion}

\newpage
\input{0appendix}

{\small
\bibliographystyle{ieee_fullname}
\bibliography{0main}
}

\end{document}

%% file: 0abstract.tex
\begin{abstract}
We show that unlike machine learning classifiers, there are no complex boundary structures in the decision boundaries for well-trained deep models. However, we found that the complicated structures do appear in training but they vanish shortly after shaping. This is a pessimistic news if one seeks to capture different levels of complexity in the decision boundary for understanding generalization, which works well in machine learning. Nonetheless, we found that the decision boundaries of predecessor models on the training data are reflective of the final model's generalization. We show how to use the predecessor decision boundaries for studying the generalization of deep models. 
We have three major findings. One is on the strength of the first principle component of deep models, another about the singularity of optimizers, and the other on the effects of the skip connections in ResNets. Code is at \url{https://github.com/hengshu1/decision_boundary_github}.
\end{abstract}



%% file: 0introduction.tex
\section{Introduction}
Decision boundary is very useful for understanding the generalization of machine learning classifiers. A decision boundary that can explain the generalization of deep neural networks is an attractive and long-standing challenge. Many efforts are related to this problem \citep{db_adversarial,db_characterizing,db_analysis,db_understanding,db_tropical,db_chen2020roby,db_hold,db_li2020adversarial,db_liu2021comprehensive,db_choi2021qimera,db_somepalli2022can,db_ouriha2022decision,db_narang2020overparameterized, db_cao2021high}.
We are going to review and discuss in details later in Section \ref{sec:related}.
In a nutshell, most works rely on the adversarial samples to characterize the decision boundary structures of deep neural networks. 

\begin{figure}[t]
\centering
\includegraphics[width=\columnwidth]{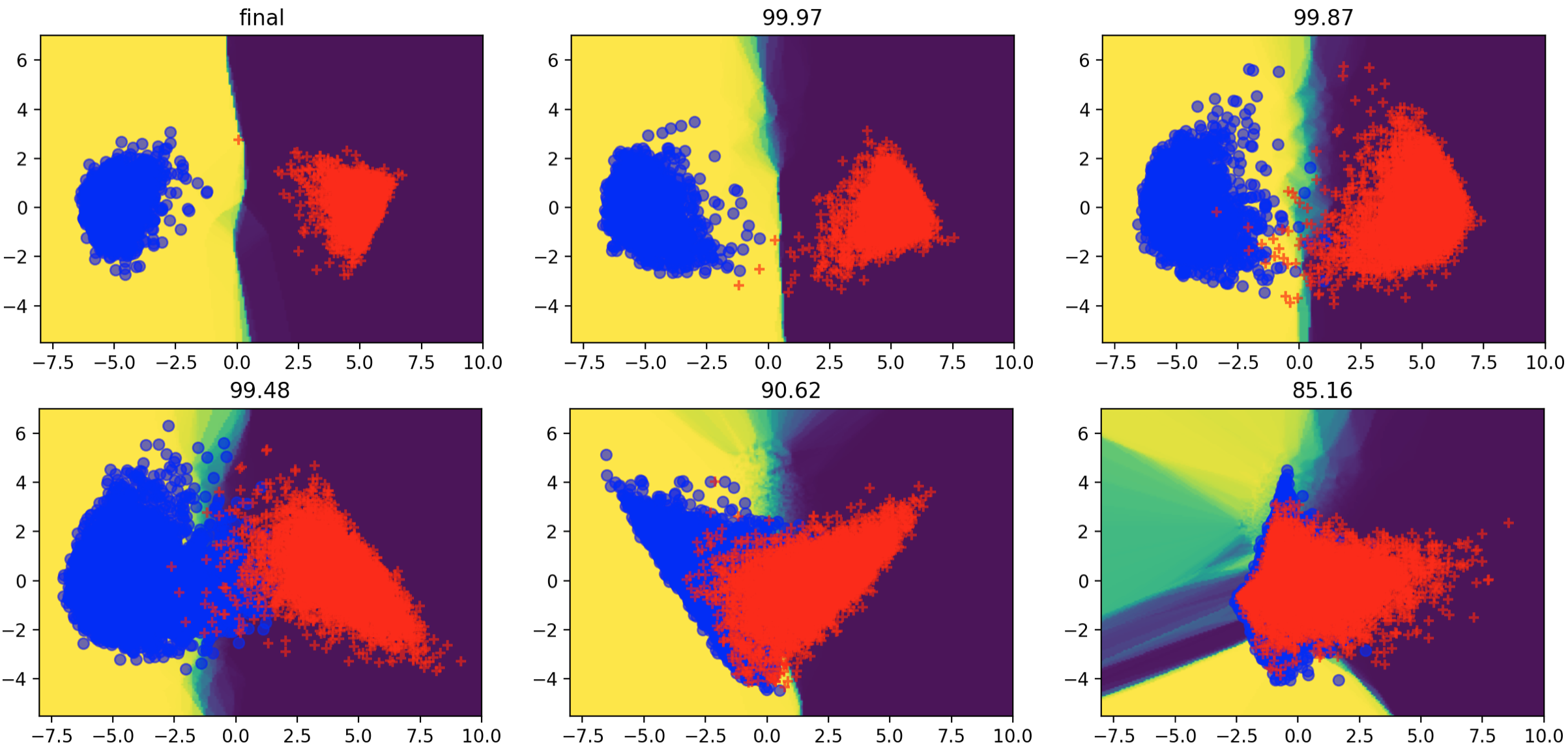}
\caption{
The decision boundary undergoes an evolutionary process. The boundaries are plotted for the final model and a number of predecessor models, using the first two principle components found by running PCA over the embedding features of all the training {\color{blue} cats} and {\color{red} dogs}. 
The heat map is plotted by first building a mapping from PCA(2) to the embedding space. See the flowchart in equation \ref{eq:flowchart}. The  Yellow/Dark-purple corresponds to the probability of CAT/DOG close to one.
Model: VGG19. 
Optimizer: SGD-anneal-lr.
}\label{fig:boundary_vanishing_pca_5}
\end{figure}

We aim to understand the decision boundary on the training samples. Not only is our method much simpler in methodology and computation, but also understanding the test performance of classifiers from the training samples is a fundamental problem. We found that, surprisingly, the decision boundaries of well trained deep models are approximately {\em linear}. It is well known that for the decision boundaries of machine learning classifiers, their complexity grows as fitting more closely to the training data. However, such complex structures do not exist for well trained deep models. The first plot of Figure \ref{fig:boundary_vanishing_pca_5} shows the case for VGG19 \citep{VGG}. \footnote{Following \citep{db_somepalli2022can}, we use the CIFAR-10 data set \citep{cifar-dataset} to visualize the decision boundary. Li et. al. \citep{NEURIPS2018_a41b3bb3} also used this data set to understand generalization by visualization (via loss contour plots).} Near the end of training, the decision boundary is indeed shaping but the complex boundary structures vanish shortly after they are formed. Interestingly, the {\em predecessor models} (especially those achieving over 99.0\% training accuracy) demonstrate an evolutionary process in the decision boundary, as shown by the rest of the plots. 
\footnote{A video of the full evolution of the decision boundary is provided in the supplementary materials.}
The boundary complexity vanishing behavior has not been observed before in both deep learning and machine learning to the best of our knowledge. 

The major contributions of this paper are as follows. 

\begin{itemize}
    
    \item Our results show that the predecessor decision boundaries on the training set are indicative of the final model's generalization performance. Visually, the models that are more self-centered, compact and have few overlapping samples near the boundaries generalize better. 
    
\item We found that deep networks decide {\em only} with the first principle component in the end. The first principle component (out of hundreds of PCA components) grows stronger and stronger during training. Eventually, the first component itself supports a wholly 100\% training accuracy. This indicates that the first principle component is associated with overfitting. 

\item There is a singularity phenomenon for optimizers. We observed that just a little reduction in the rank of an auto-correlation matrix leads to a significant generalization improvement when comparing deep learning optimizers. For training the same model, the optimizers with a low rank of this matrix generalize better than the optimizers with full-rank ones.

\item The explained variance of the first component is much larger for VGG19 than for ResNet18 and DLA. This shows the skip connections (used in both ResNets and DLA) have an effect of balancing the dominance of the first component to reduce the variances in label prediction. 
Furthermore, VGG19's auto-correlation matrix is singular, while for ResNets and DLA, the matrix is full rank.  
This shows the skip connections effectively use all the feature dimensions and extract more linearly independent features. 
    
\end{itemize}
There are other interesting observations in this paper. In particular, there is a trade-off between cluster size and class splitting for the learning rate of the SGD optimizer. 
Adam has similar variances in the first two principle directions to SGD with small learning rates. This explains why Adam has an inferior generalization to SGD with the annealed learning rate. 


%% file: 0db_evolve.tex
\section{Decision Boundary Evolution}\label{sec:evolution}
\subsection{The Decision Boundary Complexity Disappears}
Let us start with how Figure \ref{fig:boundary_vanishing_pca_5} is plotted. The VGG19 model \citep{VGG} is trained with 200 epochs of SGD and a batch size of 128. The learning rate starts with 0.1 and decays according to the Cosine rule \citep{cosine_lr_annealing}. This leads to models that are very well trained. The training accuracy is 100\% and the test accuracy is always above 93\% across many runs in our experiments. For VGG19, the CAT class has the lowest test accuracy, about 87\%, while the highest of the other classes is about 96\%. For ResNet18, CAT is 91.5\% while the best of the other classes is about 97\%. DOG is the second most mis-classified class on the test set. The mis-classification rates between CAT and DOG are the highest among all the class pairs, e.g., see \citep{cls_inter}. It thus makes sense to focus on the boundary between CAT and DOG.

Given a final and well-trained model, Principle Component Analysis (PCA) \citep{pca_wold1987principal,pca_abdi2010principal} is performed on the joint feature matrix of cats and dogs. The feature matrix is generated using the embedding features from the last layer of VGG19. We visualize the training cats and dogs according to their two major components in the first plot of Figure \ref{fig:boundary_vanishing_pca_5}. To understand how the decision changes in this space, we need to know the likelihood of any point in this space being CAT or DOG. 

Let PCA(2) $\subset \real^2$ be the space of the first two principle components. 
Denote a vector in PCA(2) by $x$. We generate a feature vector $\psi(x) \in \real^{d}$ such that it is the mean of the nearest neighbors of $x$ in PCA(2):
\[
\psi(x) = \frac{1}{n}\sum_{i=1}^{n} \phi(x_i(x)), \quad x_i(x) \in \mbox{PCA(2)},
\]
where $\{x_i(x)\}$ are the nearest neighboring training samples of $x$ in the PCA(2) space. Note that $\phi(\cdot)$ is an inverse mapping of the dimension reduction by PCA. For the training samples $\{x_i\}$, their embedding feature vectors can be queried from the networks. We take their original feature vectors for the mapping of $\phi(\cdot)$, instead of the results from the inverse transform by PCA which have large approximation errors instead. In this way, we build a mapping  $F$ that generalizes $\phi$. In particular, $F:$ PCA(2) $\to \real^{d}$, which is able to map any $x$ in PCA(2) back to the embedding space. 

The dimension $d$ is just that of the embedding space. For example, $d$ is 512 for VGG19. This gives a way of querying the neural networks model for any point in PCA(2). This process can be summarized by the flow
\begin{equation}\label{eq:flowchart}
x \stackrel{F}{\to} \psi(x) \to classifier(\psi(x)),
\end{equation}
where the last step is simply performed by a linear operation, $\psi(x) W^T +b$. Here $W,b$ are the parameters of the last layer. The softmax operation over CAT and DOG is also performed in the $classifier$ function. 
The probability of CAT is thus computed for each point in PCA(2). These probabilities generate a heat map, which is plotted in the same space with the training samples. This finishes the process of generating the first plot of Figure \ref{fig:boundary_vanishing_pca_5}. The rest of the plots in the figure are produced in the same way for some immature models during training. We call them the {\em predecessor models} in the view that they are earlier states of the final model.   

There are a few observations of this figure. For the final model (100\% training accuracy), there is no complex structure in the decision boundary. Instead, the boundary is approximately linear.
    Going back in training, we can see that there are some blurry regions in the decision boundaries of the predecessor models. These regions correspond to where the decision is ambiguous. Especially note that they are located near where the cats and dogs overlap with each other. 
Even though the overall training accuracy is descent (e.g., 99.48\%), there are still a significant number of dogs in the CAT area,  and vice versa. 
        The objects of the same classes become more and more centered and their spread size has a decreasing trend. This is especially true when training is near the end (e.g., over 99\% training accuracy) in our observation.


The first observation is quite surprising given what we know for machine learning classifiers. Usually classifiers that overfit have complex boundary structures. \footnote{See  \url{https://en.wikipedia.org/wiki/Overfitting\#/media/File:Overfitting.svg} for an illustration of how overfitting contributes to the complexity of the decision boundary in machine learning.} How come the well-trained deep models do not have such complexity in their decision boundaries? It did not make sense to us in the beginning. We further plotted the cats and dogs in the decision space, which is shown by Figure \ref{fig:boundary_vanishing} (see Appendix \ref{sec:db_in_decision}).
This confirms the decision boundary undergoes an evolution (in both the PCA space and the decision space), and finally the boundary has no complicated structures indeed.

\subsection{Predecessor Boundaries are Indicative}\label{sec:pre_indicative}
We have shown so far that the final model has no complex decision boundary structures but during training they do appear when training is near the end. Why is it important? We found that the decision boundaries of the predecessor models are indicative of the final model's generalization performance.   

\begin{figure}[t]
\centering
\includegraphics[width=\columnwidth]{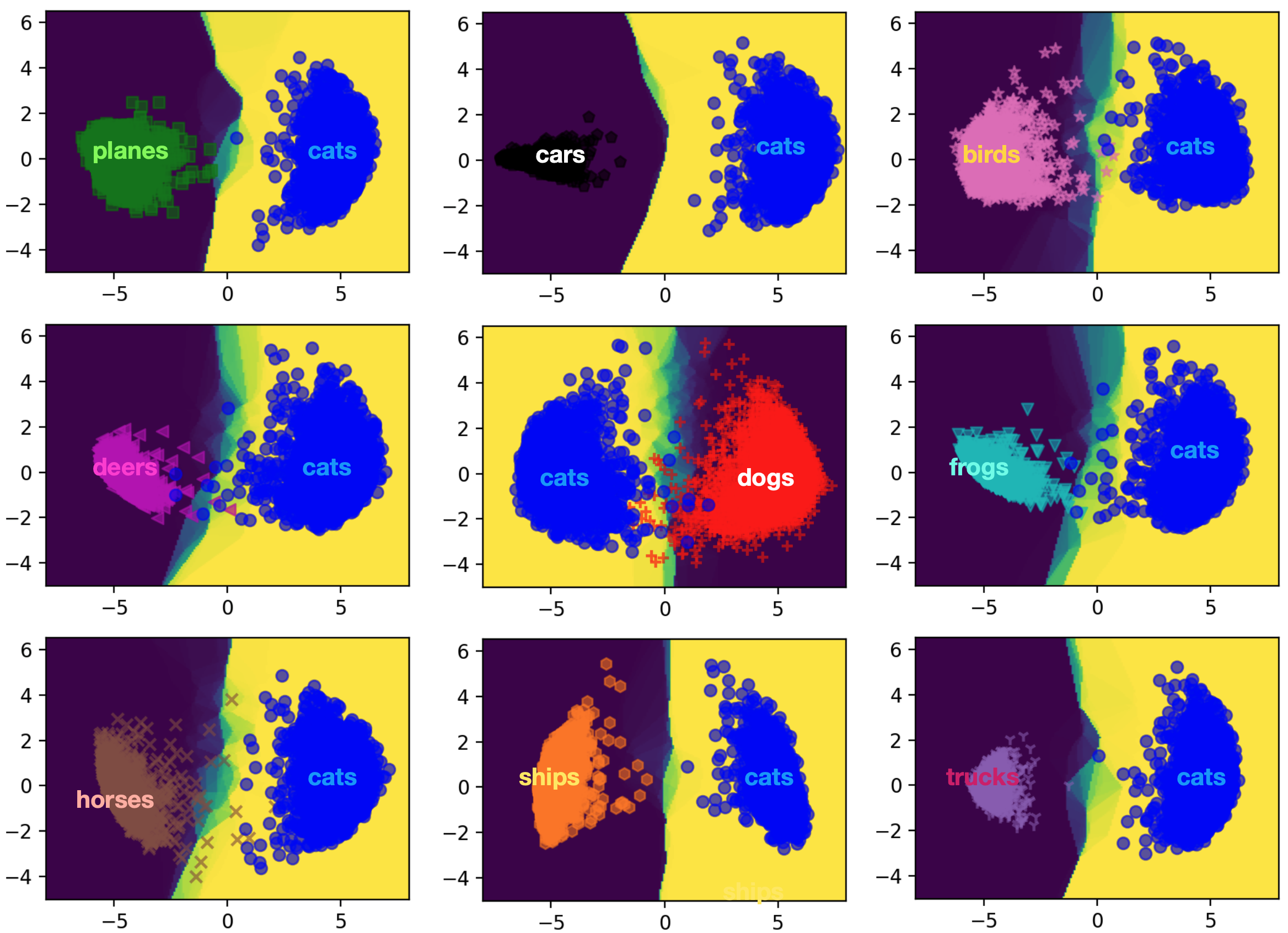}
\caption{Decision boundaries between CAT and all the other classes. This is plotted using the 99.87\%-model (in Figure \ref{fig:boundary_vanishing_pca_5}) and all the training objects.  Model: VGG19. Optimizer: SGD-anneal-lr. 
See the text in Section \ref{sec:pre_indicative} for details.
}\label{fig:boundary_cats_and_others}
\end{figure}

Figure \ref{fig:boundary_cats_and_others} shows the decision boundaries of CAT versus all the other classes in the PCA(2) space. For each class pair, a joint feature matrix is formed by querying the network's embedding feature output using the training samples of the two classes and PCA is performed afterwards. The model is the 99.87\%  predecessor model, which is the same one as used in Figure \ref{fig:boundary_vanishing_pca_5} and Figure \ref{fig:boundary_vanishing}. 
Previously \citep{cls_inter} showed that CAT and DOG have the poorest generalization on the test set. Here this figure shows the CAT-DOG decision boundary is the most crowded. In particular, this plot shows that there are many cats near the boundary (the more cats in an area the more solid is the blue color because the color is plotted using alpha equal to 0.6 transparency). Especially note that many cats cross into the area of DOG. The CAT-DOG plot in Figure \ref{fig:boundary_vanishing_pca_5} (the last plot in the first row; dogs are in the foreground instead there) shows there are a significant number of dogs that cross the boundary with CAT too. 


All the other classes have some sort of boundary with CAT. In particular, CAT has a significant number of samples near the boundary with BIRD (fluffy vs. feather), DEER (four-leg), HORSE (four-leg), and FROG (pointy ears vs. eyes atop). This explains their interference on the test set \citep{cls_inter} hereby using the training data. 
    
The above method can be extended to compare the decision boundaries of the classes in a 3D space. This is shown in Appendix \ref{sec:db_3d}.


\subsection{Rethinking the Decision Boundary for DL}\label{sec:rethinking}
One myth in deep learning is why deep neural networks overfit with such a level of over-parameterization but they still generalize so much better than machine learning classifiers. This seems to contradict with the common belief in machine learning that over-parameterized models are usually more complex and they easily catch the noises in data, leading to a low bias but a high variance and thus a poor generalization. This dilemma seems odd indeed. See \citep{db_somepalli2022can} for a similar feel. 

Belkin et. al. \citep{db_Belkin_2019} has a nice study on the bias-variance trade-off especially for deep learning. 
They uncovered a {\em double descent} phenomenon for deep learning. The classical bias-variance trade-off for machine learning is described by the well-known ``U'' shape of the test risk in terms of the capacity of the model. 
Belkin et. al.'s discovery is a double-``U'' shape. Following the first ``U'' shape, as the capacity of the model keeps growing into over-parameterization, the test risk deceases again due to the combined effect of a decreasing variance and a low bias. Somepalli et. al. \citep{db_somepalli2022can} confirmed this finding with a study of the width parameter of network layers. They also studied in particular the transition between under- and over-parameterized models. Their finding is that the ``instabilities'' in the decision boundary are the main reason of double descent. 
Note, though, the decision boundary they used is in the space of individual samples, which is different from ours. See Section \ref{sec:related} for detailed discussions. 

Our results indicate that the standard understanding of decision boundary in machine learning explains deep neural networks with some differences. \footnote{This is implied by the double descent phenomenon too.} Recall our results show that the final decision boundary is approximately linear and there is no complex boundary structures. Linear decision boundary is the simplest and it generalizes well with good features.
Perceptrons \citep{perceptron_mcculloch1943logical,perception_minsky2017perceptrons}, 
Support Vector Machine \citep{svm_Vapnik_95,svm_ls_suykens1999least}, Kernel methods \citep{vc_boser1992training,shawe2004kernel,hofmann2008kernel} and Logistic Regression \citep{menard2002applied,hastie2009elements} all greatly advanced Artificial Intelligence and some of them are still widely used in practice. Multi-layered Perceptrons \citep{rumelhart1986learning}, LSTM \citep{ltm_gers2000learning} and convolution neural networks \citep{krizhevsky2017imagenet,VGG,resnet,goodfellow2016deep} all use linear operation as an elementary composition at layers. Before the recent success of deep reinforcement learning, reinforcement learning algorithms with linear function approximation have supported the field for decades \citep{watkins1992q,tsi_td,bertsekas1996neuro,lagoudakis2003least,sutton2018reinforcement,szepesvari2010algorithms,bertsekas2012dynamic}, and it has shown excellent generalization with sparse and CMAC encoded features \citep{sutton_tile}. Silver's Ph.D thesis is built on Computer Go programs from linear value function approximation \citep{silver2009reinforcement}. Some program evaluates the Go game board from a linear combination of millions of binary features \citep{silver2007reinforcement}. There is nothing wrong with linear methods across fields and in fact they generalize very well. Their limitation is the requirement of good features.

For deep neural networks, the linearity of the final decision boundary is surprising given the huge number of parameters. We think this is the most fascinating part of deep neural networks: the excessive over-parameterization does not lead to highly complex decision boundary structures. 
This finding suggests that deep neural networks are structured in a way that enables learning very good features who finally land in a linear space.

For the final model, the large margins between classes like CAT and DOG indicate that all the classes are {\em linearly separable} in both the embedding space and the decision space thanks to the deeply learned features. Thus the decision boundary of the final well-trained model has a very simple linear structure. This can be observed from the final model in Figure \ref{fig:boundary_vanishing_pca_5} and in Figure \ref{fig:boundary_vanishing} for the CAT-DOG plot, and the 99.87\% model for the CAT-PLANE, CAT-CAR and CAT-SHIP in Figure \ref{fig:boundary_cats_and_others}. All the class pairs \footnote{Other class pairs are not shown because their boundaries look more clean.} can be easily separated by a straight line in the PCA(2) space for the corresponding model. 

\begin{center}
\fbox{\begin{minipage}{22em}
\centering
DL decision boundary != ML decision boundary:\\
Overfitting in deep learning has a very different effect on the decision boundary from machine learning.
\end{minipage}}
\end{center}
Overfitting in machine learning leads to complex decision boundaries. As the complexity grows to some level, the generalization of machine learning classifiers deteriorates. However, our results show that the decision boundary of the final well-trained model by deep learning is linear in the embedding space. In our opinion this is one reason that deep neural networks generalize much better than machine learning classifiers in many application areas. In fact, ``overfitting'' in deep learning has a beneficial effect for generalization as discussed in Section \ref{sec:opt_gen}.


It is common to examine the final models for machine learning classifiers once training finishes. 
However, the decision boundary of the final well-trained deep models cannot explain class interference \citep{cls_inter}, e.g., 
why CAT and DOG generalize much more poorly than the other classes even though they are also linearly separable in the PCA(2) space. Our main result is that the  predecessor boundaries are indicative of the final model's generalization. The slowness in carving a clear separation between classes in training is strongly correlated with the poor generalization between them at test time. We will dig deeper into this slowness remark when studying the auto-correlation matrix in Section \ref{sec:opt_gen}.

%% file: 0insights.tex
\section{Insights into Optimizer Generalization}
\subsection{From the Predecessor Boundary}\label{sec:boundary_insights}
Creating algorithms and models that generalize better is the heartbeat of deep learning. We show that comparing the predecessor decision boundaries on the training data gives us insights into their generalization performance. 
This section compares the effects of optimizers on models, and the next section is focused on the architectures. 

We start with comparing models trained with different optimizers.
{\em SGD-anneal-lr}. The learning rate starts with an initial value of 0.1, and then decays according to a Cosine rule. This is the optimizer used for the results presented in Section \ref{sec:evolution}. The momentum rate is 0.9, and the weight decay is $0.0005$. The same momentum rate and weight decay are also used in the following optimizers. For {\em SGD-big-lr} and {\em SGD-small-lr}, the learning rate is 0.01 and 0.0001, respectively. Adam \citep{kingma2017adam}: with default parameters in PyTorch. 
The model generating the results presented in this section is VGG19.

\begin{figure}[t]
\centering
\includegraphics[width=\columnwidth]{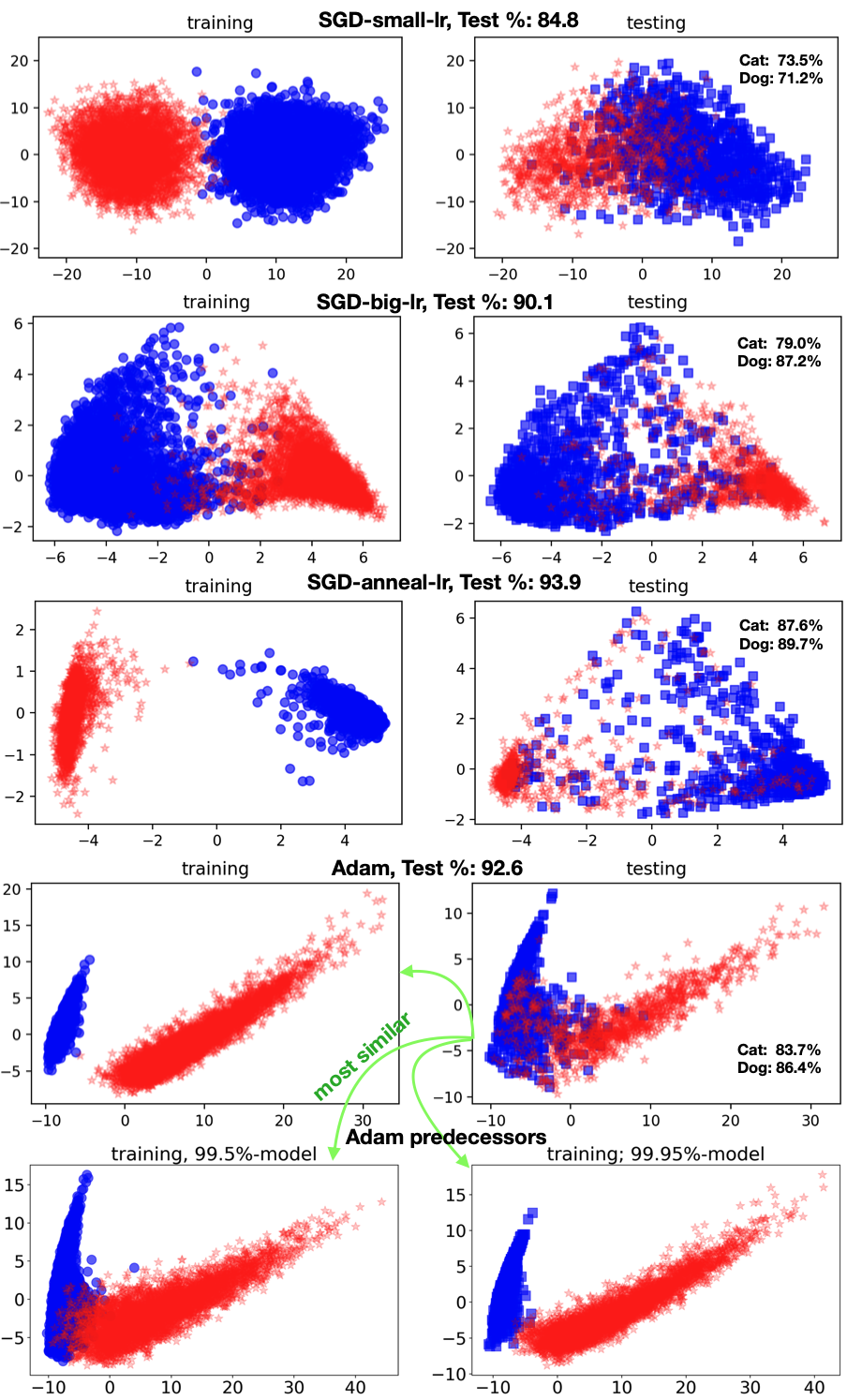}
\caption{
This shows the effect of the learning rate for SGD, in shaping the decision boundary between CAT and DOG. The Adam optimizer is also compared.
Model: VGG19. 
See the text in Section \ref{sec:boundary_insights} for details. 
}\label{fig:lr_uncover}
\end{figure}

Figure \ref{fig:lr_uncover} shows that the effect of the learning rate on the decision boundary. The overall test accuracy and the recall rates of CAT and DOG are shown in the plots as well. All optimizers are trained 200 epochs.

First, let's take a look at the plots using the training data in the first column. Small learning rates are very good at splitting classes with clean boundaries. However, the resulting cluster per class is much larger in size than using the big learning rate. Note the different ranges in the $x, y$ axes. Though, for the big learning rate, the splitting is poorer. The annealed learning rate leads to much smaller clusters with a good splitting and an even much better generalization. 
Note the large CAT areas behind the foreground DOG in the second-row plots.
It is well known that small learning rates generalize poorly in deep learning. The reason was previously explained by the sharpness of the minima \citep{flat_minima_97,sharp_minima,sharp_minima2,sharpness_yoshua,minima_valley,cls_inter}. 
This is the first time uncovering the effect of the  learning rate on the cluster size. 

Why is a big cluster bad? A big cluster due to small learning rates means the variances of the training samples are high (along the first two principle components). Intuitively, a big class cluster means that there are lots of gaps among the training samples and on the edges. Such gaps are areas where there are no training samples and thus the uncertainty is high there. At testing time, the samples are likely to slip into these gaps, which incur high variances in the label predictions. To validate this, we generate the same plot using the test data, shown at the second column of the figure (except the last row, which will be detailed later). For SGD-small-lr, there are many test samples on and across the decision boundary, which is very different from the plot on the training set. The color plot is generated using the transparency parameter alpha equal to 0.2 for DOG (foreground) and 0.6 for CAT (background). So intense red color means there are lots of dogs for the areas. The big-lr optimized model, instead, has much fewer dogs (and cats) near the test boundary. The SGD-anneal-lr optimizer further gathers most test cats and dogs in two corners, leading to the highest recall rates for the test cats and dogs as well as the overall test accuracy. 


The Adam optimizer has an interesting boundary structure. In particular, the spread range of the training cats (in blue) is small in the $x$-axis (the first component direction), similar to the annealed learning rate. Note the different ranges in the $x$-axis. However, the spread of both cats and dogs in the $y$-axis and the spread of dogs in the $x$-axis are both much wider, in fact, similar to SGD-small-lr. In addition, the boundaries of all the SGD optimizers are roughly aligned with the axes. However, for Adam, the training samples are instead rotated. There are also lots of overlapping samples near the boundary for the test data. These observations help understand why the final model by Adam is inferior to SGD-anneal-lr in generalization. 

Finally, the last row shows two predecessor models of Adam. This shows that the predecessor models on the training data are reflective of the decision boundary on the test data for Adam optimizer too. 
First, as shown by the last second row, the final model has very clear boundary structures on the training data, which is quite different from on the test data. Second, the testing plot is more similar to the training plot of the 99.5\%-model than to that of the final model. This means predecessor models give us more clues about generalization than the final model, which is consistent with the case of the SGD optimizer. This is another verification that the predecessor models are indicative of generalization. 



%% file: 0optimizer_sigma1.tex
\subsection{The Singularity of Optimizers}
\label{sec:opt_gen}
In Section \ref{sec:boundary_insights}, we plotted the predecessor boundaries for understanding the generalization capabilities of SGD and Adam optimizers. This section presents deeper insights into the generalization of the optimizers in terms of spectral properties. This draws on the ``slowness remark'' at the end of Section \ref{sec:rethinking}. Quantifying the spectral properties of the last layer gives us insights into how fast the training has converged. It appears that, as we will show later, the rate of convergence in training is strongly correlated with the generalization performance.

\begin{figure}[t]
\centering
\includegraphics[width=\columnwidth]{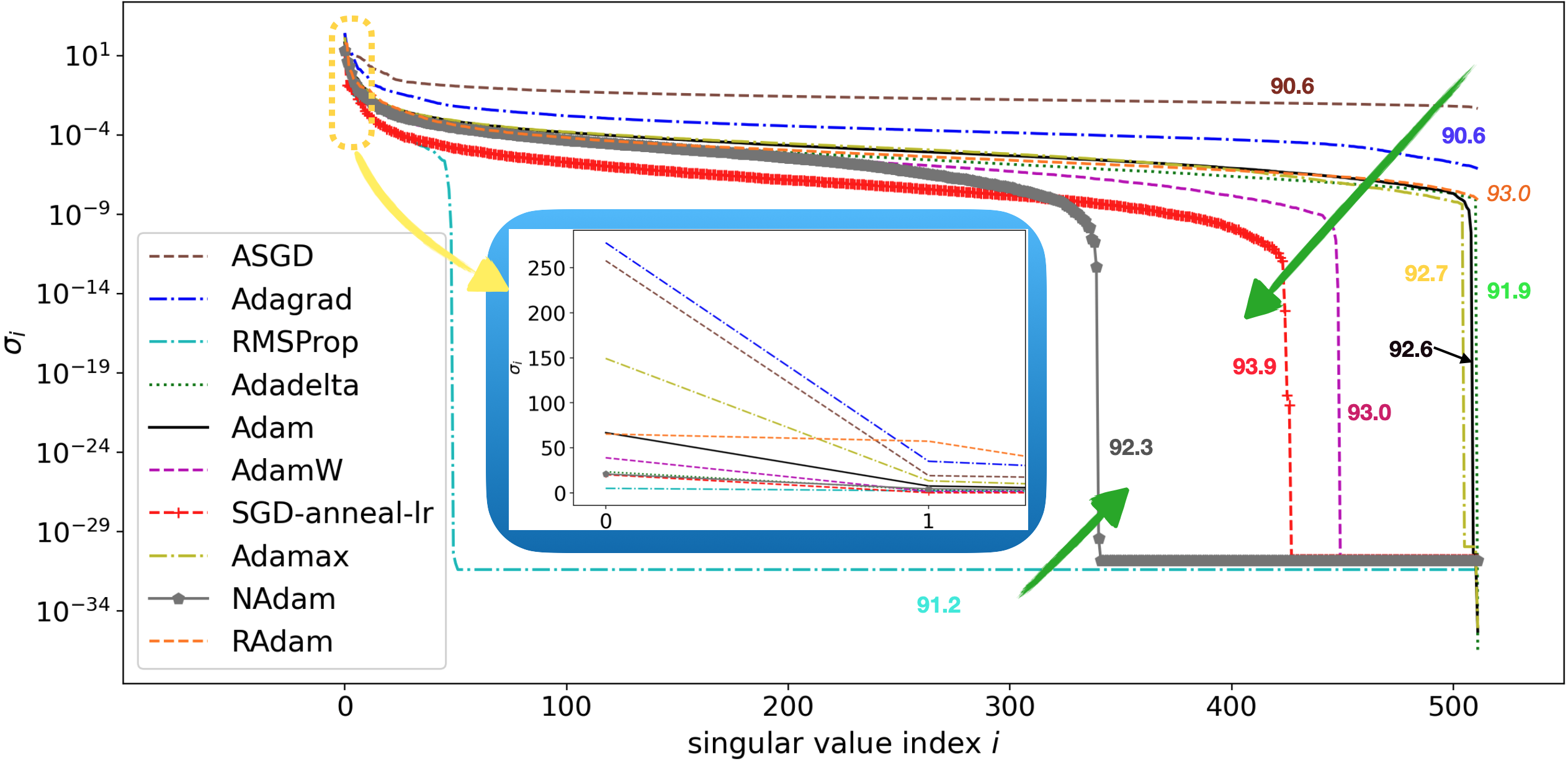}
\caption{
Optimizer profiling in terms of generalization and the spectrum of singular values of the auto-correlation matrix. The colored text number is the test accuracy of a VGG19 model trained by a corresponding optimizer. 
The snapshot in the middle shows the first two singular values.
See the text in Section \ref{sec:opt_gen} for the description of the ``singularity'' phenomenon and other observations.  
}\label{fig:opt_profile_gen_singular}
\end{figure}

Let a feature matrix be $\Phi \in \real^{n,d}$, where $n$ and $d$ are the numbers of samples and embedding features, respectively. The squared singular values of $\Phi$ are just the eigenvalues of matrix $A= \Phi^T\Phi$. This matrix $A$ is fairly important. In fact, $A$ is often called the auto-correlation matrix, which is often used for studying the convergence rate of SGD with linear function approximation in neural networks (known as the Widrow-Hoff or the Delta rule) \citep{widrow1960adaptive} and signal processing (known as the least-mean-squares filter) \citep{ljung1998system}.

Optimizing the networks up to convergence reveals the generalization capabilities of the optimizers because at convergence the data is also overfit. Thus we cover all the  optimizers (in {\em torch.optim}) that can be trained to converge by learning rate annealing. Optimizers that can not be well trained may be at a random state even training finishes. For example, SGD-big-lr is in an oscillation state at the end of training. Taking the auto-correlation matrix of such models thus compares different states of the optimizers, which is not good and thus it was avoided by us at first.

We train the same model (VGG19) using different optimizers. Each optimizer is applied with learning rate annealing in 200 epochs with a mini-batch size of 128. Figure \ref{fig:opt_profile_gen_singular} profiles ten optimizers in terms of the final test accuracy. For each optimizer, we also plot the spectrum of the singular values of the auto-correlation matrix for the CAT and DOG samples in the training set. The majority of the optimizers are on the right side. RMSprop \citep{rmsprop} and NAdam \citep{nadam} (combining Nesterov accelerated gradient \citep{nesterov} with Adam) are on the left side. If we follow the green arrows from both sides, there is a trend of generalization improving whilst the rank of this matrix gets close to some intermediate number. 
(We will discuss the RAdam optimizer later).

There is an interesting singularity phenomenon. Reducing the rank just a little (by Adadelta \citep{adadelta} and Adam \citep{kingma2017adam}) leads to a significant generalization improvement (91.9\% and 92.6\%) over the full-rank cases (ASGD and Adagrad, both 90.6\% test accuracy). AdamW \citep{adamw} and the SGD-anneal-lr further reduce the rank, which leads to more generalization improvement. The RMSProp optimizer reduces the rank to less than 10\% of the feature dimension and yet the generalization is descent. 
A large first singular value ($\sigma_1$) is a strong indicator of poor generalization, e.g., see ASGD \citep{polyak_averaging}, Adagrad \citep{adagrad}, as shown by the small blue window that sits in the middle of the plot.

\begin{figure}[t]
\centering
\includegraphics[width=0.92\columnwidth]{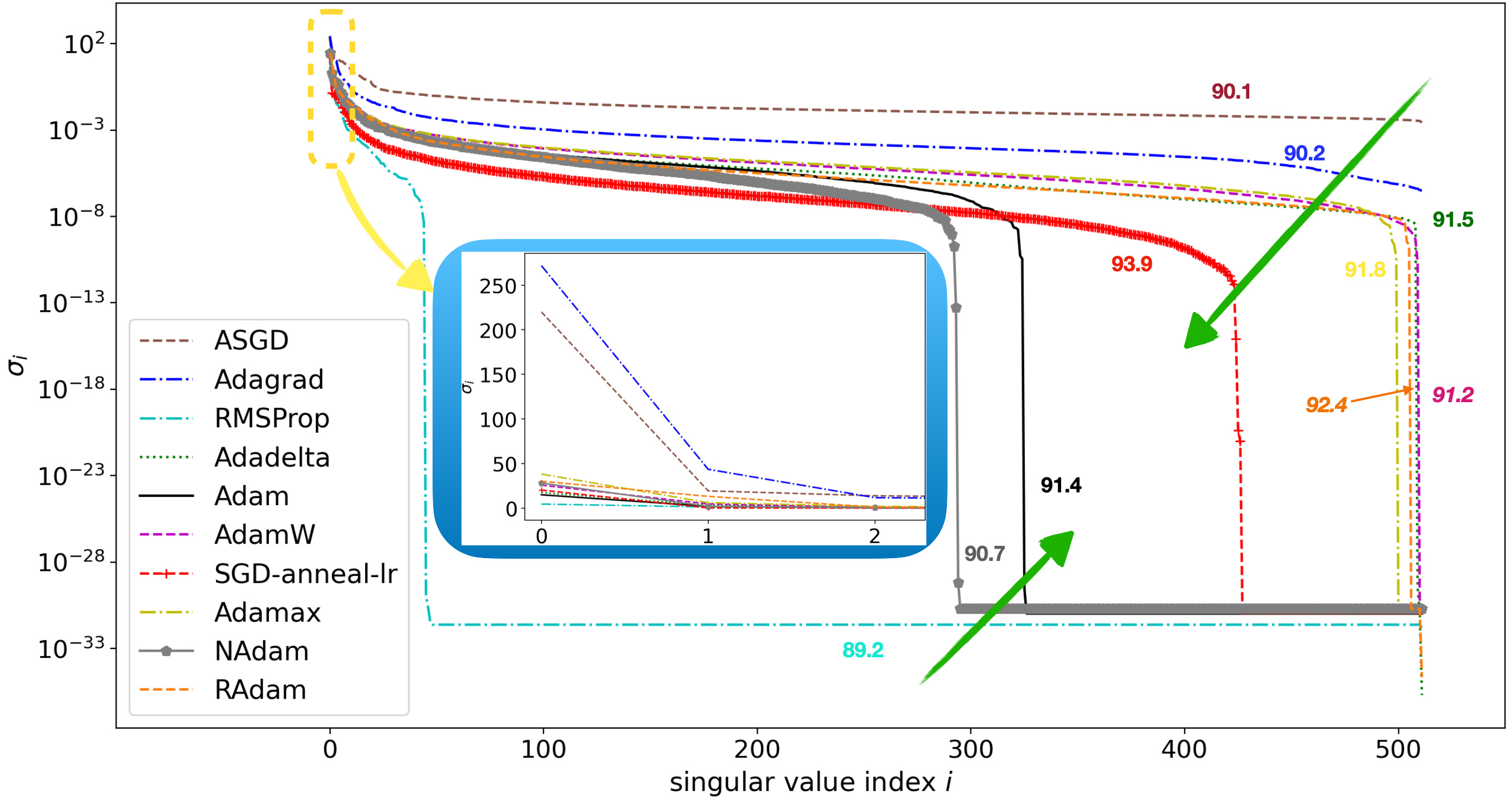}
\caption{
Similar to Figure \ref{fig:opt_profile_gen_singular}, except all the non-SGD optimizers are their default settings without learning rate annealing. Every optimizer generalizes worse than their counterpart in Figure \ref{fig:opt_profile_gen_singular}. Note the optimizers in Figure \ref{fig:opt_profile_gen_singular} are aided with learning rate annealing and they have higher training accuracies (all above 99.9\%). Thus this shows that all the (non-SGD) optimizers generalize better when the data is overfit.
The observations for Figure \ref{fig:opt_profile_gen_singular} follow similarly for the green arrow trend and the singularity of optimizers. 
}\label{fig:opt_profile_gen_singular_no_lr_anneal}
\end{figure}

In general, this figure shows when comparing different optimizers on the same model architecture, the rank of the auto-correlation matrix of the most interfering class pair is indicative of the generalization performance, and so is the $\sigma_1$. In fact, for SGD-anneal-lr, AdamW, NAdam and RMSprop, the tailing singular values are all very close to zero. This suggests these optimizers automatically filters out a significant number of noisy components in the embedding space. This is interesting because in machine learning, it is well known that dimension reduction captures major features in the data \citep{dr_carreira1997review,dr_nn_schittenkopf1997two,dr_fodor2002survey,hastie2009elements,dr_liu2016overfitting}, which leads to models that are more robust with better generalization. 

We noted that the RAdam optimizer \citep{radam} is an exception. It has a full-rank auto-correlation matrix and yet the generalization is very good (93\%). We think this is due to that the first and second singular values are of similar magnitudes, and the second principle component helps with the generalization. Out of the ten compared optimizers, RAdam is the only one that has large and close first two singular values. Adamax  \citep{kingma2017adam} is also more balanced between the first and second singular values than ASGD and Adagrad and generalizes better than them.  
 
We did run another experiment of training all the non-SGD optimizers without learning rate annealing. None of them  is able to train to 100\% accuracy. Their training accuracies are typically between 99.3\% to 99.7\%, except Adagrad reaches 99.91\%. This means all of them are not overfit comparing to SGD-lr-anneal (100\% training accuracy). Their test accuracy is shown in Figure \ref{fig:opt_profile_gen_singular_no_lr_anneal}. Surprisingly, the overfitting SGD-lr-anneal model beats all these less overfitting optimizers in generalization. 

Note that Figure \ref{fig:opt_profile_gen_singular} has learning rate annealing applied to all the non-SGD optimizers. They typically reach a training accuracy higher than 99.99\%, which is very much overfitting. Comparing the test accuracy numbers across the two figures, we can see that for each of the non-SGD optimizers, the overfitting implementation (Figure \ref{fig:opt_profile_gen_singular}) has a better generalization than the less overfitting one (Figure \ref{fig:opt_profile_gen_singular_no_lr_anneal}).
This verifies that overfitting in deep learning is distinctive --- many optimizers generalize better when they are trained very close to 100\% accuracy. ``Overfitting'' in deep learning has a different effect from machine learning, and perhaps we need a better name for it in deep learning. See Section \ref{sec:rethinking} for more discussions about this. 

The relation of the test accuracy versus the rank of the auto-correlation matrix (as indicated by the two arrow directions) in Figure \ref{fig:opt_profile_gen_singular_no_lr_anneal} is similar to that in Figure \ref{fig:opt_profile_gen_singular}, except that there is randomness for the ordering of AdamW and RAdam due to the models are not convergent. Without learning rate annealing, Adam reduces the rank significantly. It generalizes better than ASGD and Adagrad (both are full rank). However, it is still inferior to SGD-lr-anneal, which has a larger rank. This suggests that there may be an intermediate, optimal rank number, and it's not like ``the lower the rank, the better generalization''.

%% file: 0model_compare.tex
\begin{figure}[t]
\centering
\includegraphics[width=\columnwidth]{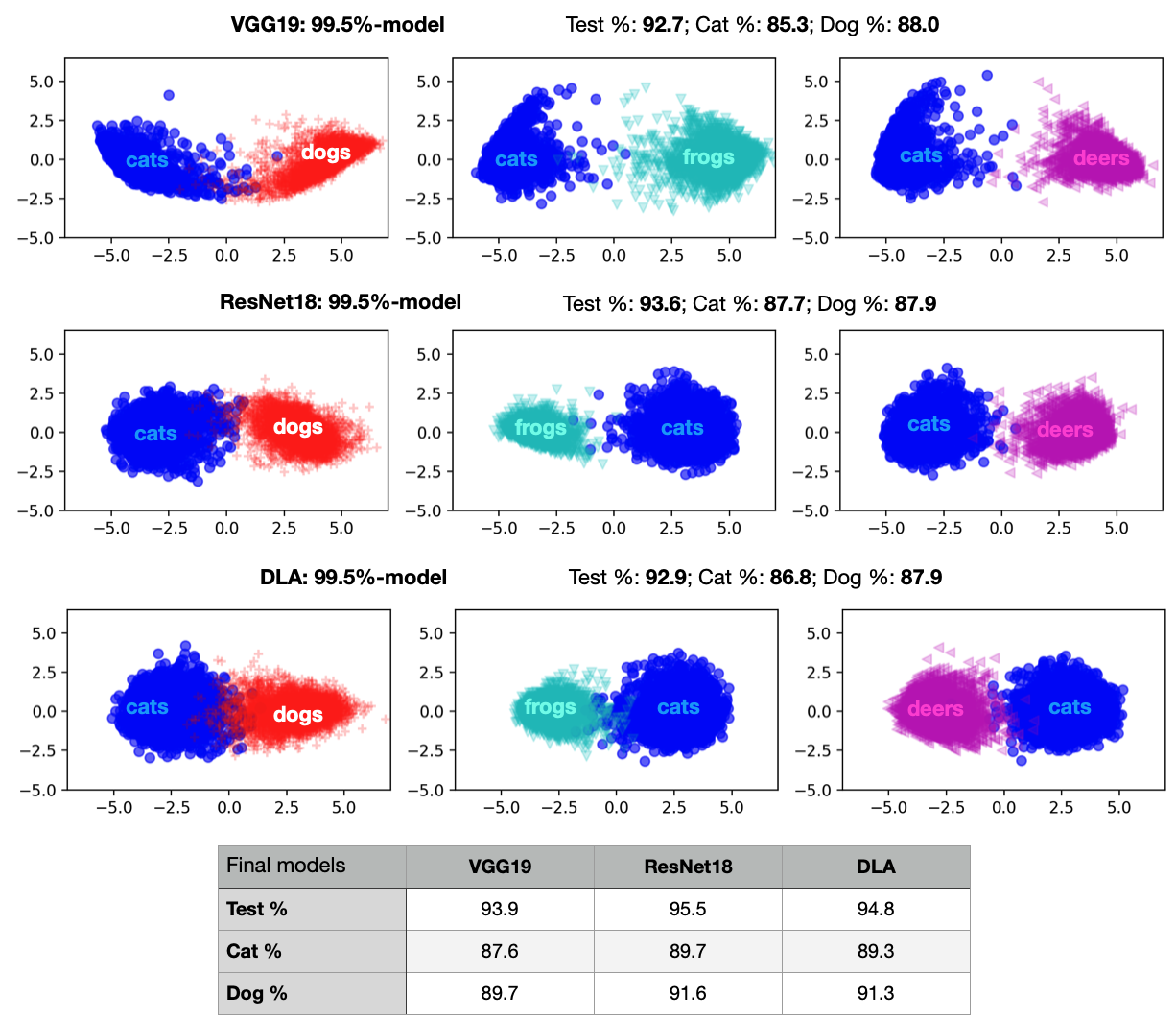}
\caption{Comparing the 99.5\% predecessor decision boundaries of VGG19, ResNet18 and DLA on the training data. For the VGG19 model, the samples are less compact, while for the ResNet18 and DLA models, the samples are distributed in a much more oval shape. The boundary complexity of ResNet18 is lower than DLA. 
}\label{fig:boundary_vgg_vs_resnet_and_dla}
\end{figure}

\section{Understanding ResNet}
The Residual network (ResNet) is an extremely popular architecture for making deep layered network models. At the time of writing, the ResNet paper \citep{resnet} received 136837 citations. The skip connection proposed in ResNet is widely adopted in designing deep models for many applications nowadays. However, why ResNets generalize well is still poorly understood. This section provides a study into this question starting from the predecessor decision boundaries. In particular, we compare VGG19, ResNet18 and DLA \citep{DLA}, optimized with SGD-anneal-lr. DLA is included because it extends ResNets into a tree structured hierarchy of layers with skip connections both within and across trees. We aim to understand the effects of the skip connections on the decision boundary and generalization.

Figure \ref{fig:boundary_vgg_vs_resnet_and_dla} shows the decision boundaries of VGG19, ResNet18 and DLA, plotted using their predecessor model at the 99.5\% training accuracy. The overall accuracy and the CAT and DOG recall rates on the test set are shown in the figure as well. The test accuracy of the final models is also summarized in the table. In terms of the generalization performance, 
ResNet18 is better than DLA, and DLA is better than VGG19. Both ResNet18 and DLA models have better self-centerness in that the samples are more clustered, while for VGG19 the samples form a less oval shape. The ResNet18 model has cleaner boundaries than DLA. 


\begin{figure}[t]
\centering
\includegraphics[width=\columnwidth]{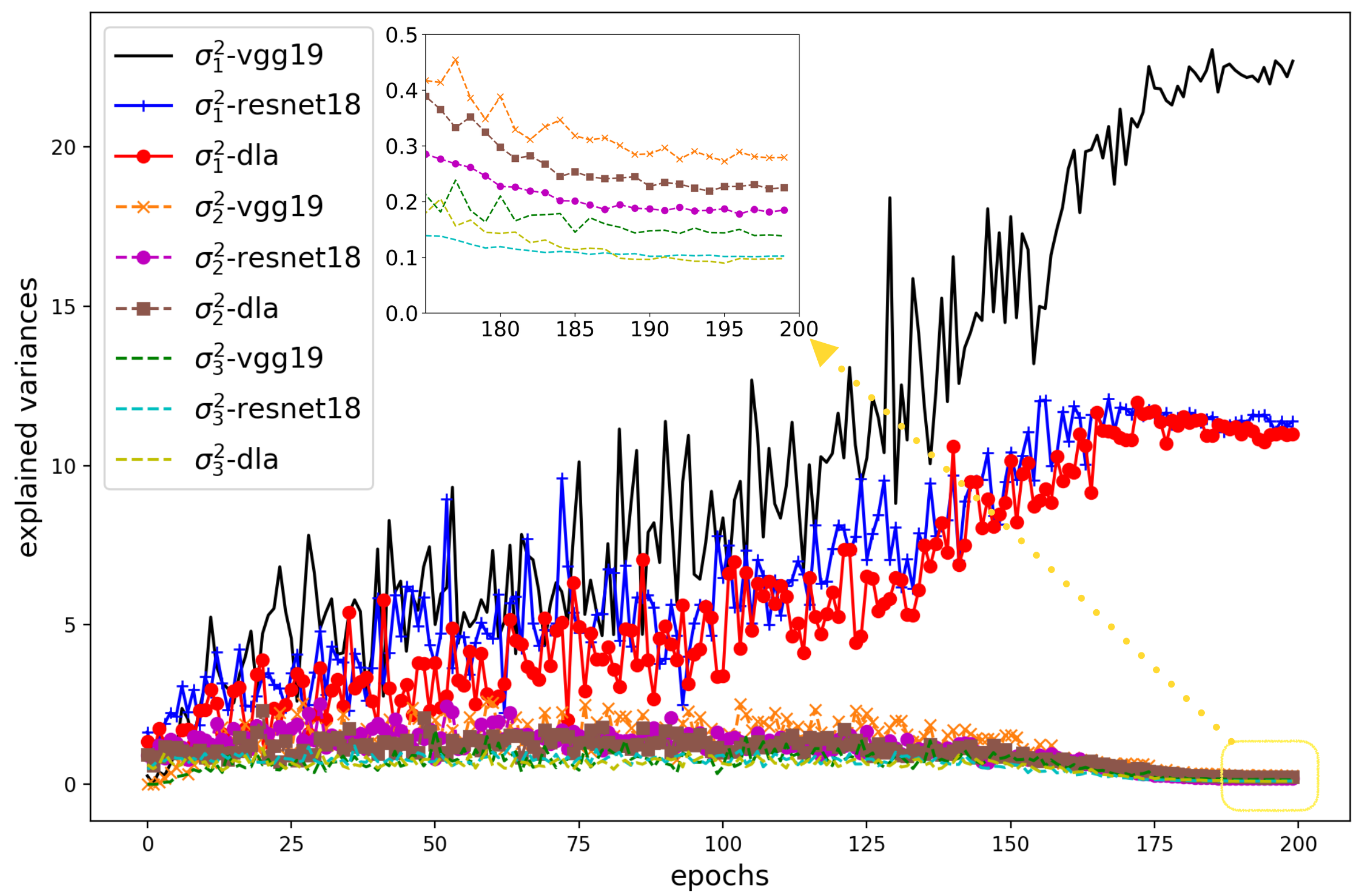}
\caption{
Evolution of the largest explained variances $\sigma_1^2$, $\sigma_2^2$ and  $\sigma_3^2$ for the training cats and dogs.
Optimizer: SGD-anneal-lr. 
}\label{fig:db_convergence_var_pca_200epochs}
\end{figure}

We were wondering why ResNets and DLA have nicer boundaries. How does the skip connection help with the decision boundary in particular? 
Figure \ref{fig:db_convergence_var_pca_200epochs} shows the evolution of the explained variance during training for VGG19, ResNet18 and DLA. The plot confirms that the first component becomes very much dominant in the end because the feature variances are mainly explained by $\sigma_1$ . 
In particular, $\sigma_1$ grows larger and larger for each model in training. Eventually the first component itself supports 100\% training accuracy for all the three models. 

VGG19 has a much bigger $\sigma_1$ than ResNet18 and DLA. The growth curve of $\sigma_1$ is also much more steep for VGG19. 
This shows that skip connections have an effect of reducing the dominance of the first component. Between ResNet18 and DLA, the explained variances are fairly close.    



\begin{figure}[t]
\centering
\includegraphics[width=\columnwidth]{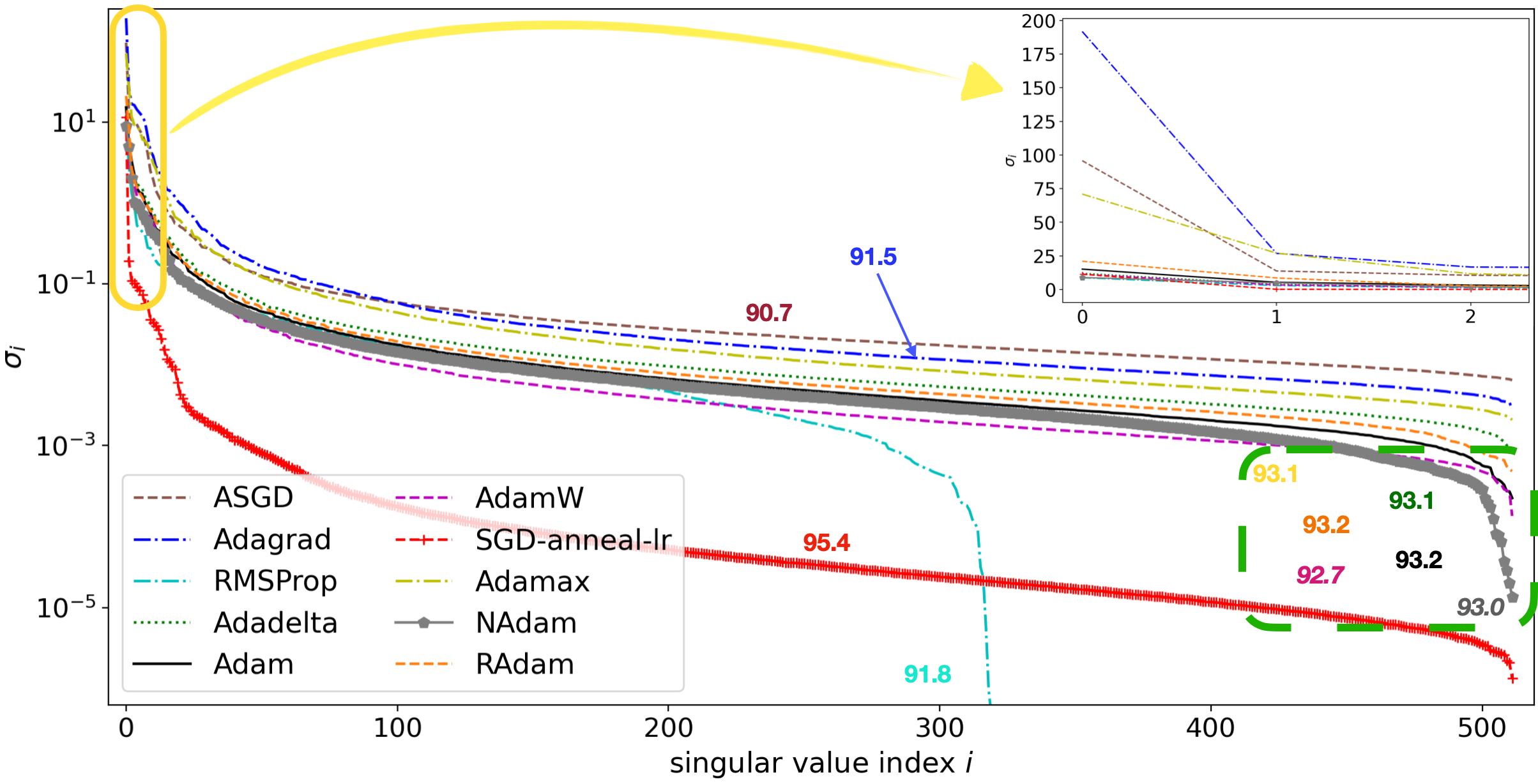}
\caption{Spectral profiling of
ResNet18 with different optimizers. One difference from VGG19 (in Figure \ref{fig:opt_profile_gen_singular}) is that here all the auto-correlation matrices are full rank. Bringing the auto-correlation matrix close to singularity in general still leads to better generalization. For example, SGD-anneal-lr continues to perform the best due to that it brings the matrix near-singular.
}\label{fig:opt_profile_gen_singular_resnet18}
\end{figure}

 Figure \ref{fig:opt_profile_gen_singular_resnet18} shows the spectral profile of ten optimizers for ResNet18. Note that for VGG19, the majority of the optimizers have a singular auto-correlation matrix, as shown in Figure \ref{fig:opt_profile_gen_singular}. However, with ResNet18, this matrix is full rank for most of the optimizers. This shows that the skip connections in ResNets have an effect of extracting more {\em linearly independent features}. This indicates that the feature dimension quota (512) is not effectively used by VGG19; and many features are linearly dependent on each other. For example, for SGD-anneal-lr, ResNet18's smallest singular value is several orders larger than VGG19. We checked the rank of the auto-correlation matrix. For VGG, it is 439 for all class pairs. For ResNet18, it is 512 (full rank) for all pairs. 
 Note this is {\em reverse} to the comparison of optimizers on the same model. In that case, reducing the rank of the auto-correlation matrix tends to improve the generalization of optimizers. However, for different architectures, it appears models that extract more linearly independent features generalize better. 
 
 About the ordering of the optimizers, in this full-rank case, though certain optimizers (ADAMW and NADAM) in the rectangle are not exactly ordered in the traversal direction of the upper green arrow in Figure \ref{fig:opt_profile_gen_singular}, their test accuracies are close. If we view the optimizers in the rectangle as a group, the green arrow traversal ordering of the generalization still holds the same as VGG19. RMSProp still outperforms ASGD and Adagrad with a much lower rank. Thus in general, for optimizing the same model, the optimizers that have a (near-)singular or low-rank auto-correlation matrix tend to generalize well. The largest singular values may play a role too. For example, the zoom view in Figure \ref{fig:opt_profile_gen_singular_resnet18} shows that ASGD and Adagrad have the top-2 $\sigma_1$s and they have the poorest generalization.  The ordering of $\sigma_1$ is not the same as the generalization performance though. Adagrad (with a larger $\sigma_1$) actually generalizes better than ASGD. Note that Adamax is more balanced between the first and second largest singular values than ASGD and Adagrad for a better generalization. This was also observed for RAdam in the VGG19 experiment. In a summary, the generalization performance of optimizers is strongly correlated with the spectral properties of this correlation matrix, and a due evaluative measure is yet to be discovered.


%% file: 0related.tex
\section{Related Work}\label{sec:related}
The popular approach in characterizing the decision boundary of deep neural networks is by adversarial attacks, which applies some perturbation to input images such that the label predictions are changed. This technique is known as Generative Adversarial Networks (GAN) \citep{gan,gan2,gan_fool}. The connection between GAN and decision boundary is that in order to generate adversarial samples, they need to cross the decision boundary. 

This fact has been used in a few works on the decision boundary of deep classifiers, e.g., see \citep{db_adversarial,db_characterizing,db_understanding,db_li2020adversarial}. 
For example, Karimi et. al. \citep{db_characterizing} generated adversarial samples, which are ambiguous to classifiers, i.e., with equal probabilities of label predictions given an object. Mickisch et. al. \citep{db_understanding} used DeepFool \citep{gan_fool} to generate adversarial samples and study the decision boundary. 
 
Guan and Loew
\citep{db_analysis} proposed a metric to evaluate the complexity of decision boundary using adversarial samples that are generated near the boundary. The key of this metric is to form a feature matrix of the adversarial samples, and then compute the Shannon Entropy of its eigenvalues.

Lei et. al. \citep{db_tao} presented a theoretical analysis on the  decision boundary (DB) variability by ($\epsilon, \eta$)-data DB  variability using a subset of training data ($\eta\%$) for a reconstruction error of $\epsilon$. They also studied the DB variability with respect to algorithms, such as training time, sample sizes, and label noise ratios. 

Somepalli et. al.
\citep{db_somepalli2022can} proposed a special decision boundary, which is defined by plotting the predicted labels of synthetic samples in a 2D plane spanned by three (real) basis images. 

The decision boundary is useful beyond understanding the generalization of deep neural networks. See Appendix \ref{sec:db_others} for other use cases.

%% file: 0conclusion.tex
\section{Conclusion}\label{sec:conclusion}
{\em The decision boundary structure is transient} for deep models. 
Our major findings are: (1) The tail of the boundary evolution is indicative of the generalization performance of deep models. (2) The first component of deep models tend to grow dominant in training and in the end, it suffices to fit the data very well by itself. (3) Optimizers have a singularity phenomenon of rank reduction in the auto-correlation matrix and improved generalization. (4) The skip connections have the effects of balancing the dominance of the first principle component and effectively using all feature dimensions. 

%% file: 0appendix.tex
\section{Appendix}

\subsection{Decision Boundary in the Decision Space}\label{sec:db_in_decision}
\begin{figure}[t]
\centering
\includegraphics[width=\columnwidth]{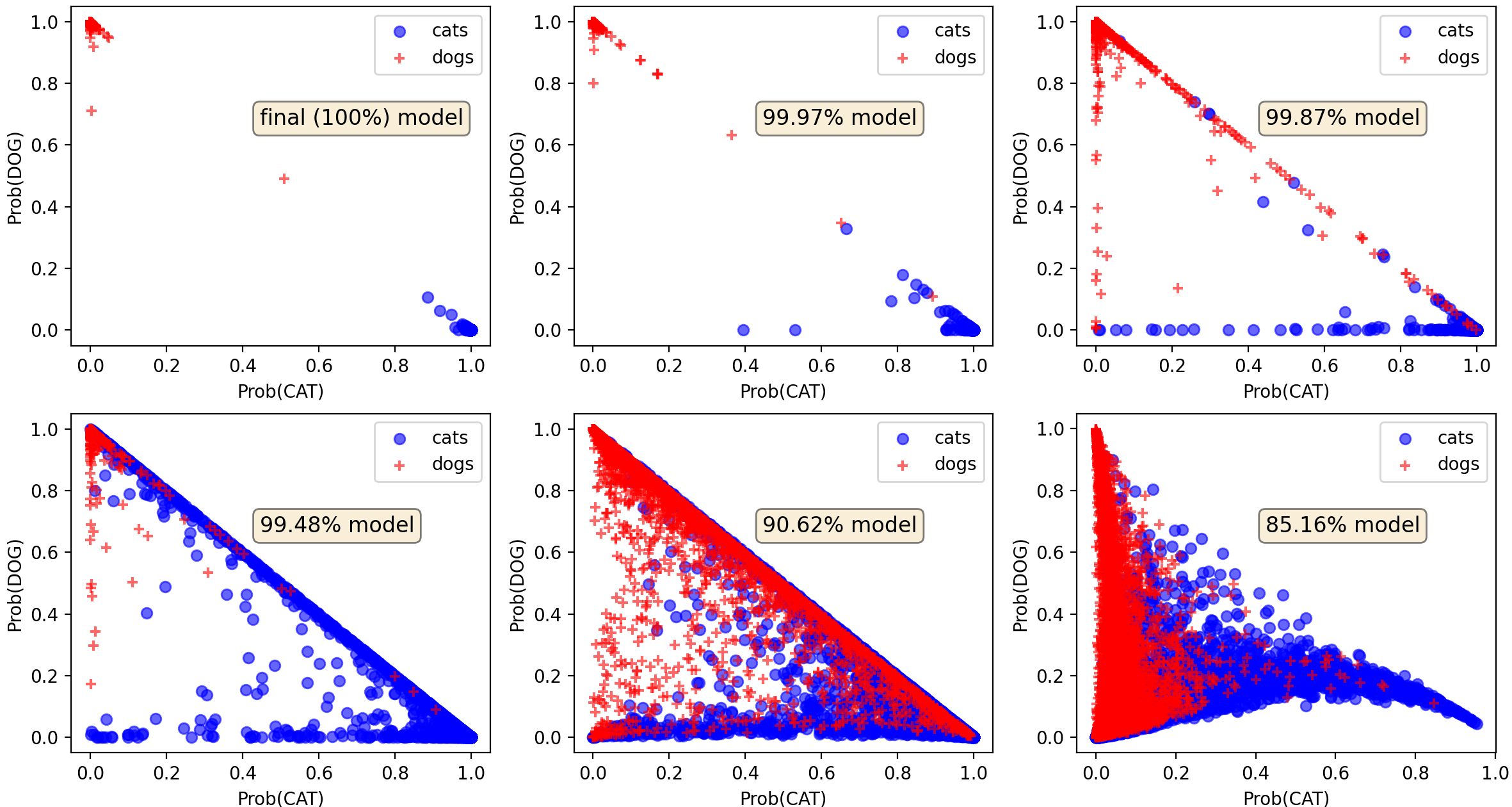}
\caption{The decision boundary vanishing phenomenon illustrated in the decision space. The $x$-$y$ plane shows the predicted probabilities of CAT and DOG for the training cats and dogs (5000 each class). The final classifier clearly separates the two classes without ambiguity and there is no complex decision boundary structure. However, the complexity does appear for the predecessor models. Model: VGG19. 
Optimizer: SGD-anneal-lr. 
}\label{fig:boundary_vanishing}
\end{figure}

The decision boundary evolution in the decision space is shown in Figure \ref{fig:boundary_vanishing}. With the final model, the samples are not only separated well but the model is very affirmative --- almost all the dogs and cats are predicted with a probability close to one. Thus studying the final model entails no decision boundary structure, and at a moment it seemed impossible to us to use the training data for studying the decision boundary. Indeed, Ortiz-Jimenez et. al. \citep{db_hold} showed that ``the decision boundaries of a neural networks can only exist when the classifier
is trained with some features that hold them together''. It is common to define the decision boundary as the set of samples for which the label prediction is equiprobable \citep{db_understanding,db_characterizing}. The problem of applying this definition is that there are no training samples that are ambiguous. 


\subsection{Decision Boundary in 3D Space}\label{sec:db_3d}
\begin{figure}[t]
\centering
\includegraphics[width=\columnwidth]{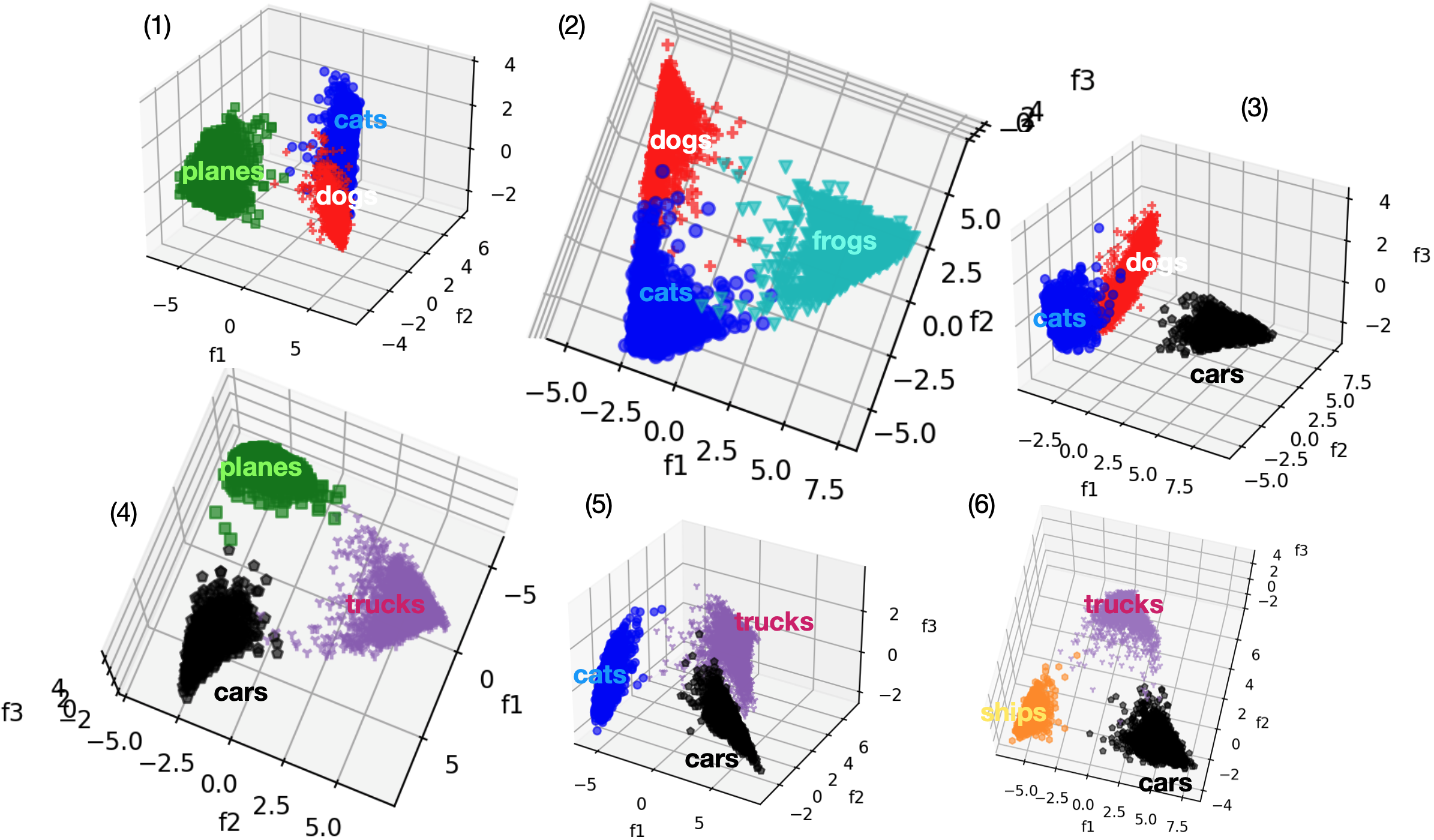}
\caption{Decision boundaries in the PCA(3) space from the feature matrix of the class triples in the training set. In particular, plot (1) shows planes are clearly separated from both cats and dogs. (2) CAT-DOG boundary is more crowded than CAT-FROG boundary, and CAT-FROG boundary is more complex than DOG-FROG boundary. (3) Cars are clearly separated from cats and dogs. (4) CAR-TRUCK is more complex than CAR-PLANE. 
(5) Cats are clearly separated from cars and trucks. (6) SHIP-TRUCK boundary is more complex than SHIP-CAR boundary, probably because SHIP is closer in size to TRUCK than CAR. These triples were chosen to show because CAR, TRUCK, PLANE, and SHIP are all metallic bodies and they interfere. These observations are consistent with the cross-class generalization between classes on the test set \citep{cls_inter}. 
}\label{fig:boundary_tripple}
\end{figure}

Visualization in 3D is convenient because it allows to zoom and rotate when inspecting the boundaries between classes. We form a feature matrix from the training objects of a class triple. Then PCA is applied to give three major components, which are the space to plot the samples. Figure \ref{fig:boundary_tripple} shows a few triples. 
The complexity of the boundaries in terms of the overlapping of training samples near the boundaries are consistent with the generalization performance on the test set.

\subsection{Use Cases of Decision Boundary}\label{sec:db_others}
Besides understanding the generalization of deep neural networks, there are also other use cases for the decision boundary. For example, Heo et. al. \citep{db_samples_distil} used  boundary samples from adversarial attacks to train a student network for the purpose of knowledge distillation.

Alfarra et al. \citep{db_tropical} found that the decision boundary of a simplest networks (affine-relu-affine) is a polytope, and showed how to use this geometric representation of the decision boundary for network pruning (i.e., reducing number of parameter by sparse matrices) and adversarial attacks. 

Choi et. al.
\citep{db_choi2021qimera} generated synthetic boundary supporting samples for the purpose of model quantization without access to the original training data.